\documentclass[10pt,twocolumn,letterpaper]{article}

\usepackage[pagenumbers]{cvpr} 

\usepackage{amsmath} 
\usepackage{enumitem} 
\usepackage{multirow}
\usepackage[table]{xcolor}
\definecolor{myLightBlue}{RGB}{240, 246, 252}

\definecolor{cvprblue}{rgb}{0.21,0.49,0.74}
\usepackage[pagebackref,breaklinks,colorlinks,allcolors=cvprblue]{hyperref}

\def\paperID{4034} 
\def\confName{CVPR}
\def\confYear{2026}

\title{Lyapunov Probes for Hallucination Detection in Large Foundation Models}

\author{
Bozhi Luan$^{1}$ \quad
Gen Li$^{1}$ \quad
Yalan Qin$^{5}$ \quad
Jifeng Guo$^{2}$ \quad
Yun Zhou$^{3}$ \quad \\
Faguo Wu$^{1}$ \quad
Hongwei Zheng$^{4}$ \quad
Wenjun Wu$^{1}$ \quad
Zhaoxin Fan$^{1,}$\thanks{Corresponding author (zhaoxinf@buaa.edu.cn).} \\
\\
{\normalsize $^{1}$Beijing Advanced Innovation Center for Future Blockchain and Privacy Computing,} \\ {\normalsize School of Artificial Intelligence, Beihang University} \\
{\normalsize $^{2}$School of Electronic and Information Engineering,}  {\normalsize State Key Laboratory of CNS/ATM, Beihang University} \\
{\normalsize $^{3}$National Key Laboratory of Information Systems Engineering, National University of Defense Technology} \\
{\normalsize $^{4}$Beijing Academy of Blockchain and Edge Computing} \\
{\normalsize $^{5}$Shanghai University}
}

\begin{document}
\maketitle

\begin{abstract}
We address hallucination detection in Large Language Models (LLMs) and Multimodal Large Language Models (MLLMs) 
by framing the problem through the lens of dynamical systems stability theory. Rather than treating hallucination as a straightforward classification task, we conceptualize (M)LLMs as dynamical systems, where factual knowledge is represented by stable equilibrium points within the representation space. Our main insight is that hallucinations tend to arise at the boundaries of knowledge—transition regions separating stable and unstable zones. To capture this phenomenon, we propose Lyapunov Probes: lightweight networks trained with derivative-based stability constraints that enforce a monotonic decay in confidence under input perturbations. By performing systematic perturbation analysis and applying a two-stage training process, these probes reliably distinguish between stable factual regions and unstable, hallucination-prone regions.
Experiments on diverse datasets and models demonstrate consistent improvements over existing baselines.

\end{abstract}

\begin{figure}[t]
  \centering
  \includegraphics[width=\linewidth]{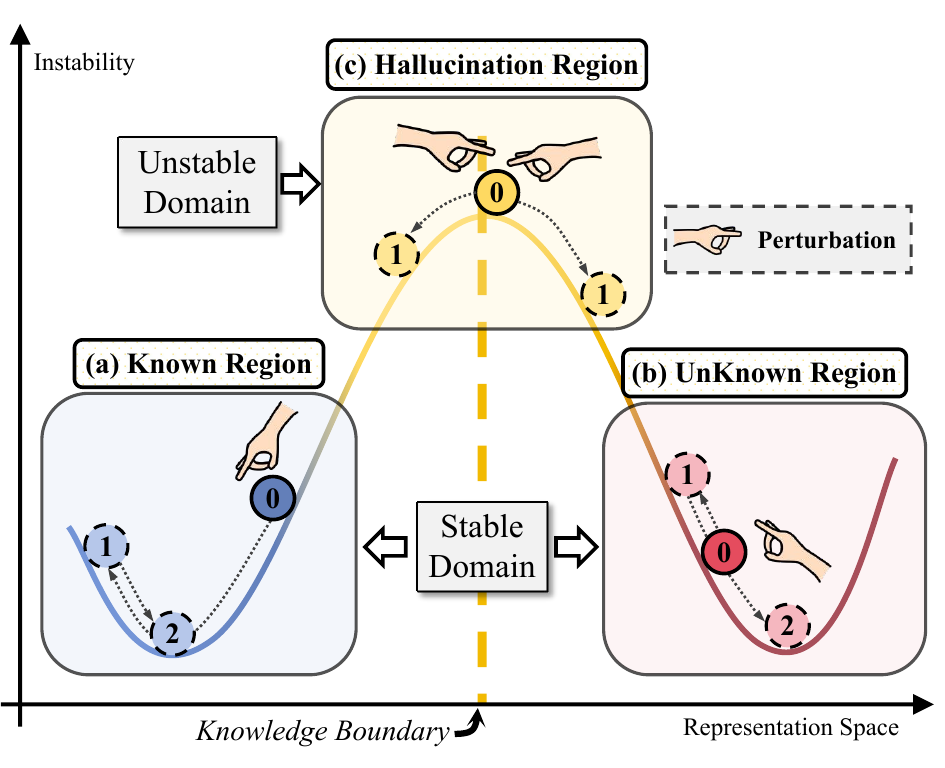}
  \caption{
   Illustration of representation space partition in large models. We define that the data (representation) space can be divided into three regions: (a) stable known regions, (b) stable unknown regions, and (c) unstable knowledge boundary regions. Hallucinations primarily emerge in the unstable boundary regions.
  }
  \label{fig:figure1}
   \vspace{-8pt}
\end{figure}

\section{Introduction}

Large Language Models (LLMs) \cite{gpt4, claude, llama, qwen} and Multimodal Large Language Models (MLLMs) \cite{gpt4v, bai2025qwenvl2.5, internvl1_5,wang2024qwenvl2} have demonstrated remarkable capabilities across diverse tasks \cite{lmmseval}, yet their tendency to generate factually incorrect content—commonly referred to as hallucinations—poses critical challenges for deployment in high-stakes domains such as healthcare, legal reasoning, and financial analysis \cite{mahaut2024factual,kadavath2022language, mllmhallucination,adaptprune,textcot}. These hallucinations manifest as plausible-sounding but factually unsupported statements, undermining trust and limiting practical applications.

Current hallucination detection approaches fall into two main paradigms: external verification methods that compare outputs against knowledge bases, and internal feature-based methods that train classifiers on model representations or token probabilities \cite{polygraph, du2024haloscope}. However, these approaches suffer from fundamental limitations. External methods require comprehensive, continuously updated fact repositories that are expensive and limited in coverage \cite{yin2023large, guo2017calibration, lin2022teaching, wang2022self, manakul2023selfcheckgpt}. Internal methods lack theoretical grounding and fail to capture the underlying mechanisms that give rise to hallucinations \cite{burns2022discovering, azaria2023internal}. Most critically, existing methods treat hallucination detection as standard binary classification without addressing the fundamental question of why and where hallucinations occur in the model's knowledge space.

We propose that the key to understanding and detecting hallucinations lies in recognizing the dynamical nature of Large Language Models and their knowledge boundaries. Our central hypothesis is that, as shown in Figure \ref{fig:figure1}, hallucinations are not randomly distributed errors but systematic phenomena concentrated at critical transition zones between regions of reliable knowledge and regions of uncertainty. These knowledge boundaries represent regions of representational instability where model behavior shifts qualitatively from fact-grounded responses to speculative generation.

To this end, drawing from dynamical system theories, we model Large Language Models \cite{bai2023qwenvl,llava,chiang2023vicuna,guo2025deepseekr1, liu2024deepseekv3} as high-dimensional dynamical systems operating in continuous representation space. In this formulation, inputs that yield accurate answers are associated with stable equilibrium points—regions where small perturbations still produce factually consistent outputs. In contrast, hallucinated content emerges near unstable points, where even minor variations can lead to significant factual deviations.  From this perspective, the definition of hallucination detection is fundamentally reframed: it becomes a matter of identifying whether a given input corresponds to a stable or unstable region in the representation space, rather than simply distinguishing between factual and non-factual outputs using arbitrary discriminative patterns.

To put this perspective into practice, we introduce Lyapunov stability theory \cite{lyapunov1992general} to analyze the robustness of language model outputs under various perturbations. Building on this foundation, we develop lightweight Lyapunov Probes that assess factual correctness by capturing the stability characteristics of different regions in the model's representation space. Our approach establishes a theoretical framework that interprets knowledge boundaries through the lens of stability transitions, providing a principled foundation for understanding where and why hallucinations emerge. The probe architecture incorporates signals from perturbation analysis and is trained with stability-oriented objectives that enforce monotonic confidence decay properties. Through carefully designed strategies, our probes effectively distinguish between stable, factual regions and unstable, hallucination-prone regions, enabling principled and efficient hallucination detection in large models.

Experiments on standard benchmarks and multiple (M)LLMs show that our method achieves consistent improvements over strong baselines. Beyond improved metrics, analysis also reveals that hallucinations consistently occur at unstable knowledge boundaries, and that stability signals are most pronounced in mid-to-late model layers across architectures. Our contribution can be summarized as:

\begin{itemize}
\item  We establish a clear connection between dynamical systems stability theory and hallucination detection, showing that knowledge boundaries can be understood as transitions between stable and unstable regions in model representations.

\item  We design Lyapunov Probes that apply stability theory in practice, using derivative-based loss functions, multi-scale perturbations, and a two-stage training process to detect hallucinations in Large Language Models.

\item We validate our approach on standard benchmarks and multiple (M)LLMs architectures, and our analysis shows that stability information is best captured in mid-to-late layers, leading to improved hallucination detection.
\end{itemize}

\section{Related Work}

\textbf{Hallucination Detection in (M)LLMs.}  
Existing hallucination detection approaches primarily rely on uncertainty scoring functions. Logit-based methods \cite{jiang2021can, yin2023large, guo2017calibration, fomicheva2020unsupervised, gal2016dropout, polygraph} assume hallucinations correspond to flat token probability distributions, while consistency-based methods \cite{wang2022self, manakul2023selfcheckgpt, kuhn2023semantic} measure agreement across multiple generated responses. Prompt-based self-evaluation approaches \cite{lin2022teaching, yin2023large, tian2023just, xiong2023can} directly query models for confidence estimates, showing effectiveness in instruction-tuned models. However, these methods typically treat hallucination detection as a pattern recognition problem and do not address the underlying reasons why hallucinations emerge at specific knowledge boundaries or how the internal representational dynamics of the model influence factual reliability. In contrast, our approach explicitly models these boundaries by leveraging stability theory, enabling a principled understanding of when and why hallucinations occur.

\begin{figure*}[t]
  \centering
  \includegraphics[width=\linewidth]{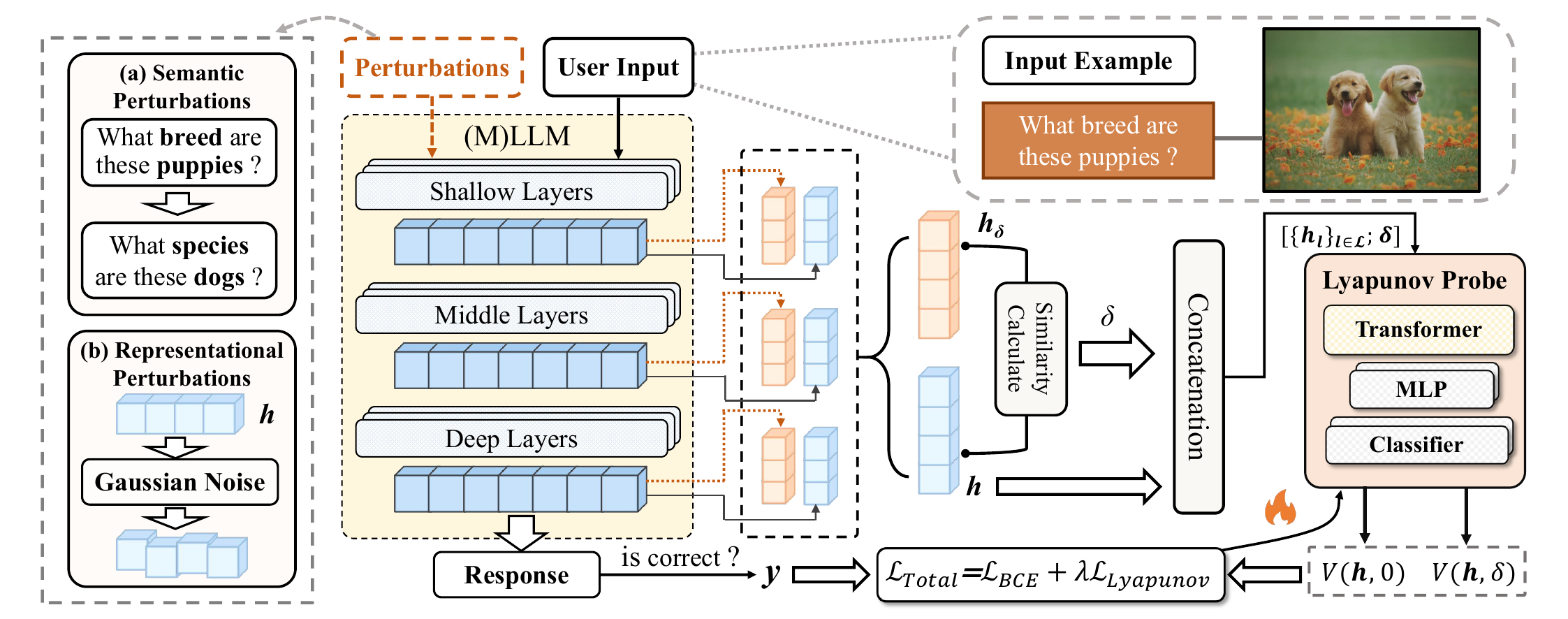}
  \caption{
     Overview of our Lyapunov Probing framework. Multi-layer hidden states and perturbation information are fed into a probe network comprising a transformer-based HiddenProcessor and an MLP-based Classifier. The framework is trained with stability-driven objectives that enforce monotonic confidence decay, enabling distinction between stable factual regions and unstable hallucination-prone regions.
  }
  \label{fig:framework}
\end{figure*}

\textbf{Probe-Based Detection Methods.}  
Recent work explores training classifiers on internal model representations to distinguish truthful from hallucinated content. Supervised probing approaches \cite{mahaut2024factual, azaria2023internal, kadavath2022language, burns2022discovering} extract confidence estimates from hidden layer activations, and multi-layered probes can capture factual knowledge encoded across different network depths. Other methods \cite{chen2024inside} perform eigen decomposition on activation covariance matrices or search for meaningful directions in representation space to identify hallucination-related patterns; for example, HaloScope \cite{du2024haloscope} estimates hallucination subspaces through explicit classifier training. While these probe-based methods effectively exploit internal representations, they mainly focus on discriminative pattern learning and lack a theoretical framework for explaining the relationship between representation stability and hallucination. Our method builds on this line by introducing Lyapunov-constrained probes, which theoretically link representational stability to factual reliability and hallucination propensity.

\textbf{Dynamical Systems and Stability in ML.}  
Neural ODEs have modeled deep networks as continuous-time dynamical systems to analyze training dynamics \cite{ode,ding2023improving, westny2023stability}, and adversarial robustness research investigates how input perturbations affect predictions \cite{cipriani2024minimax}, exposing decision boundary sensitivities. Lyapunov stability theory has been applied to study training convergence and certify robustness \cite{rodriguez2022lyanet}, but these applications mainly concern computational or adversarial stability during or after training. In contrast, our work extends dynamical systems and Lyapunov theory to the epistemic domain, using it to characterize knowledge boundaries and detect hallucinations in pre-trained (M)LLMs. This provides a new perspective on hallucination detection, grounded in the stability properties of internal representations rather than solely in output behaviors.

\section{Method}

In this section, we introduce our Lyapunov Probes for hallucination detection. Our objective is to accurately identify instances where model outputs deviate from factual knowledge and exhibit hallucinated content. Rather than relying solely on surface features or output distributions, we approach this challenge from a dynamical systems perspective, modeling the evolution of internal representations within the (M)LLMs as a high-dimensional dynamical process. This viewpoint enables us to analyze how model knowledge, uncertainty, and hallucination emerge as distinct stability properties in representation space. 

We begin by formalizing the hallucination detection task and present our dynamical modeling framework, which partitions knowledge space based on stability theory. Then, we detail the design of a Lyapunov-guided probe network and its associated training objectives, followed by our perturbation strategies and optimization procedure for enforcing stability constraints. Together, these components provide a principled and interpretable approach to detecting hallucination-prone regions in (M)LLMs. Next, we introduce each component in detail.

\subsection{Dynamical System Modeling of Large Models}
\label{sec:dynamical}

We model (M)LLMs \cite{bai2023qwenvl,llava,chiang2023vicuna}  as high-dimensional dynamical systems, where hallucination detection is reframed as the analysis of the stability properties of internal representations. Specifically, for a given input, we apply controlled perturbations to these representations and extract hidden states from chosen (M)LLMs' layers, allowing us to probe their stability in representation space. As illustrated in Figure~\ref{fig:framework}, this approach enables us to distinguish between regions associated with factual knowledge and those prone to hallucination.

Formally, we conceptualize the (M)LLMs' forward computation as a sequence of transitions governed by a dynamical system $\mathcal{F}: \mathbb{R}^d \rightarrow \mathbb{R}^d$, where $d$ is the dimensionality of the hidden states. For an input sequence $x$, the hidden state evolves across layers as:
\begin{equation}
h^{(l+1)} = \mathcal{F}^{(l)}(h^{(l)}), \quad h^{(l)} \in \mathbb{R}^d
\end{equation}

Within this dynamical view, factual knowledge is associated with \textbf{attracting regions} in the representation space—regions where small perturbations to the input or internal state result in outputs that remain factually consistent. Conversely, hallucinated content is linked to trajectories that either fail to converge or settle in \textbf{unstable regions} where minor variations can cause significant changes in the model’s response.

To clarify the geometric and behavioral structure of the model’s knowledge, we partition the representation space into three zones:
\begin{itemize}
    \item \textbf{Stable Knowledge Region} ($\mathcal{S}_\text{K}$): Contains inputs that are well-grounded in the model's factual knowledge. For $x \in \mathcal{S}_\text{K}$ with representation $h = \mathrm{Encoder}(x)$, small perturbations $\delta$ satisfy $\| \mathcal{F}(h + \delta) - \mathcal{F}(h) \| < \epsilon$ for $\|\delta\| < \epsilon_0$, indicating robust and consistent outputs.
    \item \textbf{Stable Unknown Region} ($\mathcal{S}_\text{U}$): Contains inputs outside the model’s factual scope, but where outputs remain stable even under small perturbations. That is, $\| \mathcal{F}(h + \delta) - \mathcal{F}(h) \| < \epsilon$, but the model consistently outputs “unknown” or abstains from speculation.
    \item \textbf{Unstable Knowledge Boundary Region} ($\mathcal{B}$): This transitional zone lies between the above two regions and is characterized by conditional or fragile stability. Here, the model's response to small perturbations can change abruptly, and hallucinations are most likely to occur.
\end{itemize}

To provide a principled framework for identifying and measuring these unstable regions, we employ Lyapunov stability theory. Specifically, we define a probe function $V(h, \delta)$ that estimates the probability that a given representation under perturbation remains factually correct. The Lyapunov stability condition requires that the probe’s confidence should decrease monotonically as the perturbation magnitude increases, providing a principled way to distinguish between stable factual knowledge and unstable, hallucination-prone regions.

\subsection{Lyapunov Probes Design}
\label{sec:probe}

To realize the above analysis in practice, we propose a lightweight and adaptive model that fuses multi-layer Transformer hidden states and perturbation information. The core objective of the model is to output a confidence score in the range [0, 1] (closer to 1 indicates higher factuality and a more stable state) by automatically learning stability-related features and enforcing monotonicity constraints. The probe takes both the multi-layer selected original hidden representation and explicit perturbation strength as input, specifically concatenating \(\{h_{l}\}_{l \in \mathcal{L}}\) and $\delta$. This design enables the network to capture the relationship between perturbation magnitude and output stability, with the final output given by a sigmoid activation:
\begin{equation}
V(h, \delta) =  \text{Classifier}\left( \text{HiddenProcessor} (\{h_{l}\}_{l \in \mathcal{L}}; \delta  ) \right),
\end{equation}
where we utilize a transformer to capture inter-layer dependencies through the self-attention mechanism in the HiddenProcessor component, which is followed by a 2-layer feature projector that generates task-relevant representations.
The Classifier is a simple 3-layer MLP structure, designed to output prediction confidence. The Probe is trained using a composite loss function:
\begin{equation}
\mathcal{L}_{\mathrm{total}} = \mathcal{L}_{\mathrm{BCE}} + \lambda \mathcal{L}_{\mathrm{Lyapunov}} ,
\end{equation}
where the two main objectives correspond to different Lyapunov-inspired properties.

First, the binary cross-entropy loss $\mathcal{L}_{\mathrm{BCE}}$ supervises the probe to predict factual correctness on unperturbed samples:
\begin{equation}
\mathcal{L}_{\mathrm{BCE}} = -\mathbb{E}\left[ y \log V_0 + (1-y) \log(1-V_0) \right] ,
\end{equation}
where $V_0 = V(h, 0)$ denotes the prediction on unperturbed representations, and $y \in \{0,1\}$ indicates whether the model can correctly answer the query. This loss trains the probe to assess the model's factual knowledge at stable states (without perturbation), establishing a baseline confidence aligned with the Lyapunov requirement that the function should peak at stable equilibria.

Second, the Lyapunov constraint loss $\mathcal{L}_{\mathrm{Lyapunov}}$ explicitly enforces the monotonic decay property of the Lyapunov function, requiring that the probe's confidence decreases as the perturbation magnitude increases:
\begin{equation}
\mathcal{L}_{\mathrm{Lyapunov}} = \mathbb{E}_{h, \delta} \left[ \max\left(0, \frac{\partial V(h, \delta)}{\partial \delta}\right) \right] ,
\end{equation}
where this constraint enforces the Lyapunov stability condition $\frac{\partial V(h, \delta)}{\partial \|\delta\|} < 0$, which mandates that the Lyapunov function $V$ decreases as the perturbation magnitude $\delta$ increases. By penalizing non-negative derivatives, the loss ensures that larger perturbations to the representation $h$ consistently lead to lower predicted factuality confidence, aligning with the stability requirement that deviations from stable states reduce the system's confidence in factual correctness.

Our approach leverages hidden states from three strategically selected layers to measure model confidence. Early-layer representations are rich in semantic content, middle-layer hidden states provide the strongest discriminative signal for confidence estimation, and late-layer representations increasingly reflect the model's output generation process. This multi-layer signal aggregation yields a more reliable confidence indicator than single-layer approaches.

\subsection{Perturbation and Training Strategies}
\label{sec:perturb}

A critical aspect of our approach is the design of perturbation strategies during training that effectively probe the robustness of model representations. To help the probes evaluate stability, we employ a combination of semantic and representational perturbations to comprehensively test stability:

\begin{itemize}
\item \textbf{Semantic Perturbations}: These include controlled variations such as substitution of words from the same grammatical class, insertion of random tokens, and adjustment of sentence structure. They ensure the probe learns to distinguish between cases where core factual content remains stable amid linguistic shifts and those where such variations alter the underlying truth.
\item \textbf{Representational Perturbations}: These involve direct modifications to the hidden states by injecting Gaussian noise. Such perturbations simulate small, random fluctuations in the model’s internal representations, which are designed to systematically push the representation towards and across knowledge boundaries.
\end{itemize}

\begin{table}[t]
\setlength{\tabcolsep}{0.6mm}
\centering
\small
  \caption{Main results. Comparison with competitive hallucination detection methods on different datasets across different models. We use AUPRC as the metrics.}
  \begin{tabular}{l|l|c c c c}
    \toprule
    \textbf{Model}        & \textbf{Method}    & \textbf{TriviaQA}  & \textbf{PopQA}     & \textbf{CoQA}      & \textbf{MMLU}      \\
    \midrule
    \multirow{5}{*}{Llama-2-7B}   & Verbalized \cite{tian2023just}     & 58.37     & 20.13     & 51.52     & 28.99     \\
                              & Surrogate   \cite{xiong2023can}    & 57.14     & 18.42     & 53.29     & 30.77     \\
                              & Seq. Prob. \cite{jiang2021can} & 63.76     & 20.22     & 49.63     & 29.41     \\
                              & Probe \cite{mahaut2024factual}          & 75.99     & 61.87     & 72.25     & 33.60     \\
                              & \cellcolor{myLightBlue}Ours            & \cellcolor{myLightBlue}83.09     & \cellcolor{myLightBlue}63.37     & \cellcolor{myLightBlue}76.13     & \cellcolor{myLightBlue}33.79     \\
    \midrule
    \multirow{5}{*}{Llama-3-8B}   & Verbalized  \cite{tian2023just}    & 64.72     & 21.23     & 52.40     & 54.09     \\
                              & Surrogate  \cite{xiong2023can}     & 66.02     & 18.26     & 53.50     & 52.43     \\
                              & Seq. Prob.  \cite{jiang2021can} & 70.72     & 27.02     & 50.35     & 57.48     \\
                              & Probe      \cite{mahaut2024factual}       & 78.82     & 60.77     & 80.67     & 79.26     \\
                              & \cellcolor{myLightBlue}Ours            & \cellcolor{myLightBlue}86.46     & \cellcolor{myLightBlue}67.08     & \cellcolor{myLightBlue}81.28     & \cellcolor{myLightBlue}80.00     \\
    \midrule
    \multirow{5}{*}{Qwen-3-4B}  & Verbalized \cite{tian2023just}     & 43.54     & 12.04     & 62.56     & 67.61     \\
                              & Surrogate  \cite{xiong2023can}     & 39.70     & 10.08     & 59.54     & 65.43     \\
                              & Seq. Prob.  \cite{jiang2021can} & 49.82     & 13.79     & 61.13     & 71.70     \\
                              & Probe      \cite{mahaut2024factual}       & 74.47     & 64.41     & 88.30     & 87.35     \\
                              & \cellcolor{myLightBlue}Ours            & \cellcolor{myLightBlue}79.47     & \cellcolor{myLightBlue}64.02     & \cellcolor{myLightBlue}89.01     & \cellcolor{myLightBlue}87.48     \\
    \midrule
    \multirow{5}{*}{Falcon-7B} & Verbalized \cite{tian2023just}     & 35.18     & 13.76     & 35.54     & 23.90     \\
                              & Surrogate  \cite{xiong2023can}     & 37.23     & 16.53     & 34.83     & 23.78     \\
                              & Seq. Prob.  \cite{jiang2021can} & 40.00     & 15.13     & 37.13     & 24.98     \\
                              & Probe    \cite{mahaut2024factual}         & 63.27     & 60.48     & 65.36     & 24.79     \\
                              & \cellcolor{myLightBlue}Ours            & \cellcolor{myLightBlue}65.52     & \cellcolor{myLightBlue}61.23     & \cellcolor{myLightBlue}66.03     & \cellcolor{myLightBlue}25.11     \\
    \bottomrule
  \end{tabular}
  \label{tab:main_results}
\end{table}

For each input, we construct a sequence of perturbations ${\delta_1, \ldots, \delta_K}$ with controlled, incremental magnitudes, where the intensity of both semantic and representational perturbations gradually increases. We calculate $\delta$ as the cosine similarity between the unperturbed representation $h$ and the perturbed representation $h_{\delta}$: $\delta = 1 - \cos(h, h_{\delta})$. This balances the need to make stability transitions observable while preserving the underlying semantics of the input.

Training proceeds in two stages. In the first stage, the probe is trained using binary cross-entropy loss to distinguish factual from non-factual outputs. In the second stage, the Lyapunov constraint loss is gradually introduced with increasing weight $\lambda$, enforcing monotonic confidence decay as perturbations intensify. This approach ensures stable optimization while establishing desired stability properties.

\section{Experiments}

To assess the efficacy of our proposed Lyapunov Probes, we conduct comprehensive experiments across a diverse array of LLMs and MLLMs. Our evaluation framework spans six representative models, covering a range of architectures and scales, as well as eight carefully curated benchmarks designed to evaluate distinct aspects of hallucination: factual recall, dialogue consistency, cross-domain knowledge, and multimodal grounding. Next, we detail our experiments.

\subsection{Experimental Setup}

\textbf{Models.} We evaluate our method on six open-source models: Llama-2-7B-Chat \cite{touvron2023llama2}, Llama-3-8B-Instruct \cite{grattafiori2024llama3}, Qwen-3-4B-Instruct \cite{yang2025qwen3}, and Falcon-7B-Instruct \cite{almazrouei2023falcon} for LLMs, and LLaVA-1.5-7B \cite{llava1.5} and Qwen-2.5-VL-3B \cite{bai2025qwenvl2.5} for MLLMs. All models are used in their official instruction-tuned variants, with greedy decoding to ensure deterministic outputs and fair comparisons.

\textbf{Datasets.} The datasets utilized in our evaluation encompass both language-only and multimodal hallucination scenarios, ensuring comprehensive coverage of the problem space. For LLMs, we employ TriviaQA \cite{joshi2017triviaqa} and PopQA \cite{popqa} to evaluate factual question-answering, targeting trivia knowledge and popular knowledge gaps, respectively. Multi-turn conversational consistency is assessed using CoQA \cite{reddy2019coqa}, while broad-domain multiple-choice reasoning across 57 subjects is benchmarked with MMLU \cite{mmlu}. For MLLMs, we leverage POPE \cite{pope} to examine object existence hallucinations in images, TextVQA \cite{textvqa} to measure scene-text recognition and question-answering capabilities, and VizWiz-VQA \cite{gurari2018vizwiz} to evaluate real-world visual question answering on user-generated, low-quality images. Additionally, MME \cite{mmebenchmark} is integrated as a holistic benchmark for multimodal perception and reasoning.

\textbf{Baselines.} We compare our Lyapunov Probes against four representative baseline approaches. Verbalized confidence \cite{tian2023just} uses prompt-based self-estimation, directly querying instruction-tuned models for numerical confidence scores, though it can suffer from miscalibration. Surrogate methods \cite{xiong2023can} employ a smaller auxiliary language model to compute token probabilities for target tokens, balancing probability-based objectivity with prompting accessibility. Average Sequence Probability \cite{jiang2021can} computes mean log-probability across generated tokens as a model-intrinsic confidence estimate, though it tends toward overconfidence and sensitivity to linguistic variations. The Probe baseline \cite{mahaut2024factual} trains standard supervised classifiers on hidden states without stability constraints, isolating the contribution of our Lyapunov-theoretic framework.

\begin{figure*}[t]
  \centering
  \includegraphics[width=\linewidth]{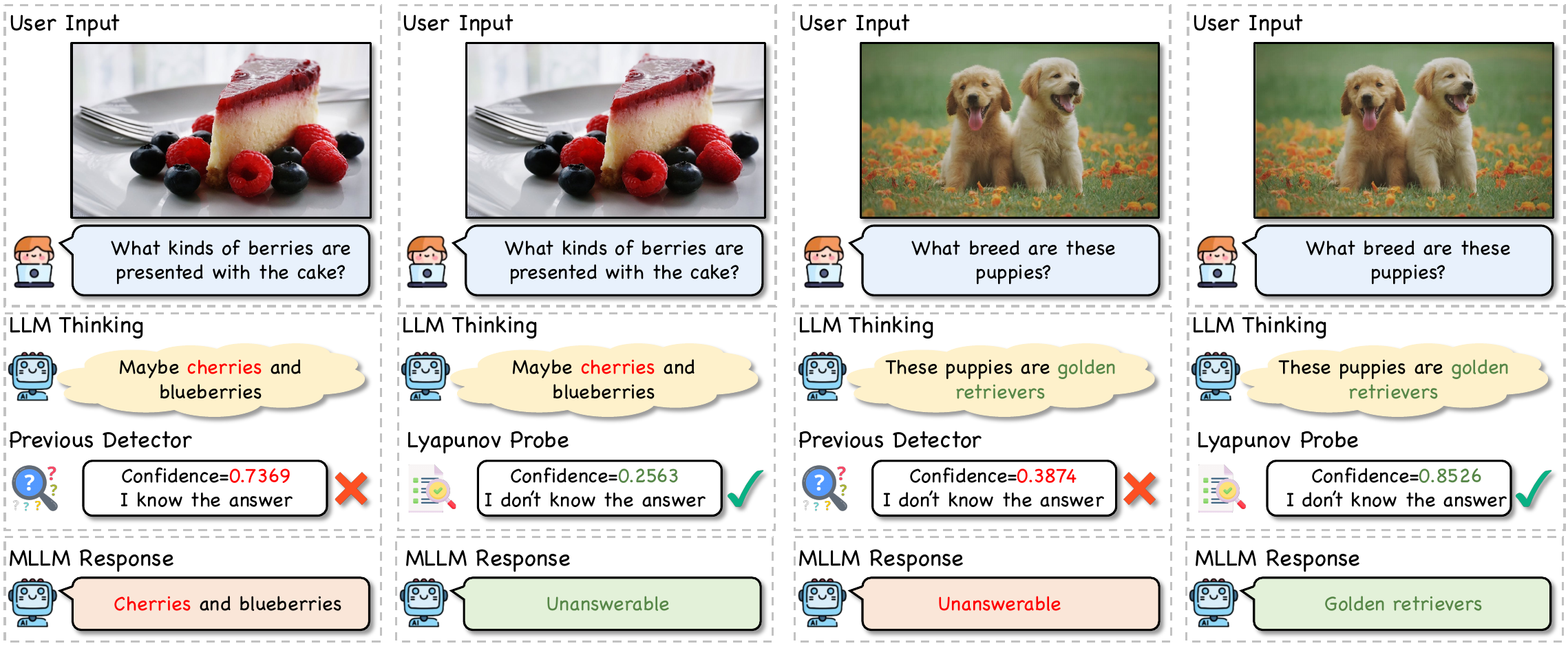}
  \caption{
     Case analysis of our Lyapunov Probe. The probe score provides an accurate prediction of whether the model can answer correctly, enabling the system to either proceed with or abstain from generating a response, thereby effectively reducing hallucinations.
  }
  \label{fig:case_1}
\end{figure*}

\textbf{Metrics.} Following \cite{mahaut2024factual, kadavath2022language}, we adopt AUPRC (Area Under Precision-Recall Curve) as our primary metric, which is particularly suited for the class imbalance inherent in hallucination detection datasets.  
By integrating precision and recall across all decision thresholds, AUPRC directly measures our goal of reliably distinguishing stable factual regions from unstable hallucination-prone regions. All experiments use greedy decoding, 80/20 train/validation splits, and multiple random seeds for statistical reliability.

\begin{table}[t]
\setlength{\tabcolsep}{1.0mm}
\centering
\small
\caption{MLLM Experiment on different datasets across different models. We use AUPRC as the metric.}
\begin{tabular}{l | l | c c c c}
    \toprule
    \textbf{Model}          & \textbf{Method}  & \textbf{POPE}  & \textbf{TextVQA} & \textbf{VizWiz} & \textbf{MME}   \\
    \midrule
    \multirow{2}{*}{LLaVA-1.5} & Probe \cite{mahaut2024factual}  & 98.08 & 85.89   & 77.02  & 93.61 \\
            & \cellcolor{myLightBlue} Ours    &\cellcolor{myLightBlue} 99.13 &\cellcolor{myLightBlue} 89.02   &\cellcolor{myLightBlue} 83.18  &\cellcolor{myLightBlue} 95.18 \\
    \midrule
    \multirow{2}{*}{Qwen-2.5-VL}  & Probe \cite{mahaut2024factual}  & 98.41 & 95.61   & 84.04  & 96.32 \\
            & \cellcolor{myLightBlue} Ours    &  \cellcolor{myLightBlue} 99.00 &  \cellcolor{myLightBlue} 96.98   & \cellcolor{myLightBlue}  85.17  &  \cellcolor{myLightBlue} 97.57 \\
\bottomrule
\end{tabular}
\label{tab:mllm_performance}
\vspace{-6pt}
\end{table}

\subsection{Results on Large Language Models}
Table~\ref{tab:main_results} presents the AUPRC results across four LLMs and four benchmarks, highlighting the consistent effectiveness of our method. Lyapunov Probes achieve an average improvement of 6.2\% over standard probes and 18.5\% over probability-based baselines. The most notable gains are observed on open-ended factual QA tasks, such as TriviaQA and PopQA, where our approach demonstrates strong performance in addressing parametric knowledge gaps. For instance, on TriviaQA, Lyapunov Probes deliver a 7.1\% improvement with Llama-3-8B.

Model-specific trends further emphasize the advantages of our method. Llama-3-8B achieves the highest overall performance with an average AUPRC of 78.7\%, making the most of its robust pretraining to effectively leverage our perturbation-based training. This is particularly evident on PopQA, where Llama-3-8B achieves a 5.3\% improvement, excelling on rare and niche entities where other methods struggle. Falcon-7B, while showing lower absolute performance at 65\%, still benefits from our framework with an average relative uplift of 2.0\%, demonstrating the versatility of our approach across different model capabilities. On the dialogue-focused CoQA benchmark, smaller average gains of 1.5\% can be attributed to the contextual coherence inherent in the task. However, Qwen-3-4B performs exceptionally well, achieving 89.01\%, thanks to its compact design and ability to handle incremental perturbations effectively.

Across all benchmarks, Lyapunov Probes consistently achieve 4-8\% improvements on tasks requiring factual accuracy. Overall, these results demonstrate the effectiveness of Lyapunov Probes in improving the robustness and reliability of Large Language Models across a variety of tasks.

\begin{table}[t]
  \centering
  \caption{Cross-domain Dataset Transfer Experiment Results. Probes are trained on the TriviaQA dataset and evaluated on the CoQA and PopQA datasets.}
  \begin{tabular}{l | l | c c}
    \toprule
    \textbf{Model}        & \textbf{Method}                  & \textbf{CoQA}      & \textbf{PopQA}     \\
    \midrule
    \multirow{5}{*}{Llama-2-7B}   & Verbalized   \cite{tian2023just}           & 51.52     & 20.13     \\
                              & Surrogate  \cite{xiong2023can}               & 53.29     & 18.42     \\
                              & Seq. Prob. \cite{jiang2021can}        & 49.63     & 20.22     \\
                              & Cross-domain Probe      & 71.47     & 54.48     \\
                              & \cellcolor{myLightBlue}In-domain Probe         & \cellcolor{myLightBlue}76.13     & \cellcolor{myLightBlue}63.37     \\
    \midrule
    \multirow{5}{*}{Llama-3-8B}   & Verbalized    \cite{tian2023just}          & 52.40     & 21.23     \\
                              & Surrogate    \cite{xiong2023can}             & 53.50     & 18.26     \\
                              & Seq. Prob.   \cite{jiang2021can}      & 50.35     & 27.02     \\
                              & Cross-domain Probe      & 67.65     & 53.47     \\
                              & \cellcolor{myLightBlue}In-domain Probe         & \cellcolor{myLightBlue}81.28     & \cellcolor{myLightBlue}67.08     \\
    \midrule
    \multirow{5}{*}{Qwen-3-4B}  & Verbalized    \cite{tian2023just}          & 62.56     & 12.04     \\
                              & Surrogate   \cite{xiong2023can}              & 59.54     & 10.08     \\
                              & Seq. Prob.   \cite{jiang2021can}      & 61.13     & 13.79     \\
                              & Cross-domain Probe      & 73.12     & 48.73     \\
                              & \cellcolor{myLightBlue} In-domain Probe         & \cellcolor{myLightBlue}89.01     & \cellcolor{myLightBlue}64.02     \\
    \midrule
    \multirow{5}{*}{Falcon-7B} & Verbalized   \cite{tian2023just}           & 35.54     & 13.76     \\
                              & Surrogate   \cite{xiong2023can}              & 34.83     & 16.53     \\
                              & Seq. Prob.  \cite{jiang2021can}       & 37.13     & 15.13     \\
                              & Cross-domain Probe      & 47.75     & 39.74     \\
                              & \cellcolor{myLightBlue}In-domain Probe         & \cellcolor{myLightBlue}66.03     & \cellcolor{myLightBlue}61.23     \\
    \bottomrule
  \end{tabular}
  \label{tab:cross_domain_transfer_full}
\end{table}

\subsection{Results on Multimodal Large Language Models}
For MLLMs, Table~\ref{tab:mllm_performance} reports AUPRC results on four vision-language benchmarks, where our method achieves an average improvement of 2.1\% over base probes. This demonstrates the capability of our approach to address multimodal misalignment effectively. By incorporating both textual and visual perturbations (e.g., image noise with $\sigma=0.1$), our method uncovers performance gaps that traditional techniques fail to address.

On POPE, near-saturation at an average AUPRC of 99.0\% indicates that the task is well-suited to existing baselines, with our method providing a modest 0.8\% improvement by refining boundary-level predictions. Larger gains appear on perception-intensive tasks: LLaVA-1.5-7B achieves a 3.2\% improvement on TextVQA, reaching 89.02\%, where challenges such as noisy fonts and OCR errors are effectively mitigated by our approach. The most significant gains are observed on VizWiz-VQA, which involves real-world, low-quality user-generated images. Here, our method achieves an average improvement of 3.6\%, with Qwen-2.5-VL-3B reaching 85.17\%. This highlights the ability of our method to handle noisy and ambiguous inputs where baseline methods struggle.

On the holistic MME benchmark, our approach achieves an average improvement of 1.4\%. Qwen-2.5-VL-3B stands out with an AUPRC of 97.57\%, demonstrating its ability to leverage balanced vision-language pretraining for enhanced performance. Notably, LLaVA sees a significant 6.2\% improvement on VizWiz-VQA, where degraded visual inputs challenge traditional approaches.

Qualitative examples in Figure~\ref{fig:case_1} further illustrate the practical benefits of our approach. For ambiguous objects and conceptually confusing cases, our Lyapunov Probes effectively mitigate both overconfidence and excessive uncertainty in challenging scenarios, yielding more accurate and reliable responses compared to baseline methods.

Overall, these results confirm that our method extends effectively to multimodal settings, delivering consistent improvements across diverse benchmarks.

\begin{table*}[t]
\setlength{\tabcolsep}{6.0mm}
\centering
\caption{Ablation study on key components. Performance drops when removing each component, demonstrating their necessity for the full method.}
\label{tab:ablation_models}
\begin{tabular}{l|cccc}
\toprule
\textbf{Settings} & \textbf{Llama-2-7B} & \textbf{Llama-3-8B} & \textbf{Qwen-3-4B} & \textbf{Falcon-7B} \\
\midrule
w/o Perturbation Data   & 82.41 & 82.35 & 79.92 & 65.65 \\
w/o Two-Stage Training & 82.00 & 84.80 & 80.58 & 64.27 \\
w/o Multi-layer Hidden States & 77.34 & 82.16 & 77.59 & 60.50 \\
w/o Lyapunov Constraint Loss & 78.13 & 82.86 & 74.19 & 62.48 \\
\rowcolor{myLightBlue} Our Full Lyapunov Probe & 83.09 & 86.46 & 79.47 & 65.52 \\
\bottomrule
\end{tabular}
\end{table*}

\section{Ablation Study}

To assess the contribution of each component within our Lyapunov Probes framework, we conduct comprehensive ablation experiments on the TriviaQA dataset. We systematically verify the theoretical soundness of our approach, demonstrate its generalization capabilities, and validate the necessity of key design choices.

\subsection{Verification of Lyapunov Stability Properties}

\begin{figure}[t]
  \centering
  \begin{subfigure}[b]{0.48\linewidth}
    \centering
    \includegraphics[width=\linewidth]{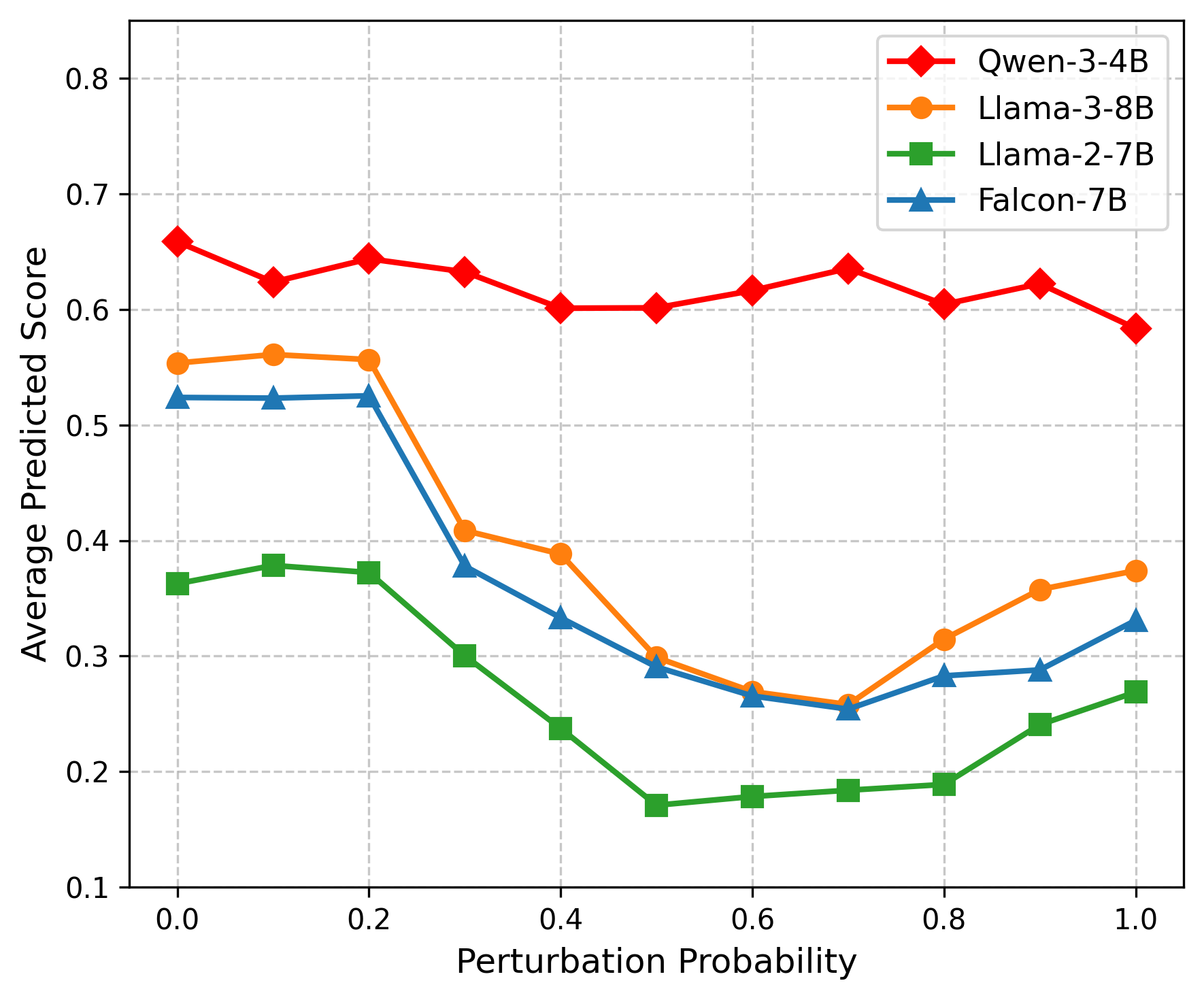}
    \caption{Base Probe}
    \label{fig:layer:1}
  \end{subfigure}
  \hfill 
  \begin{subfigure}[b]{0.48\linewidth}
    \centering
    \includegraphics[width=\linewidth]{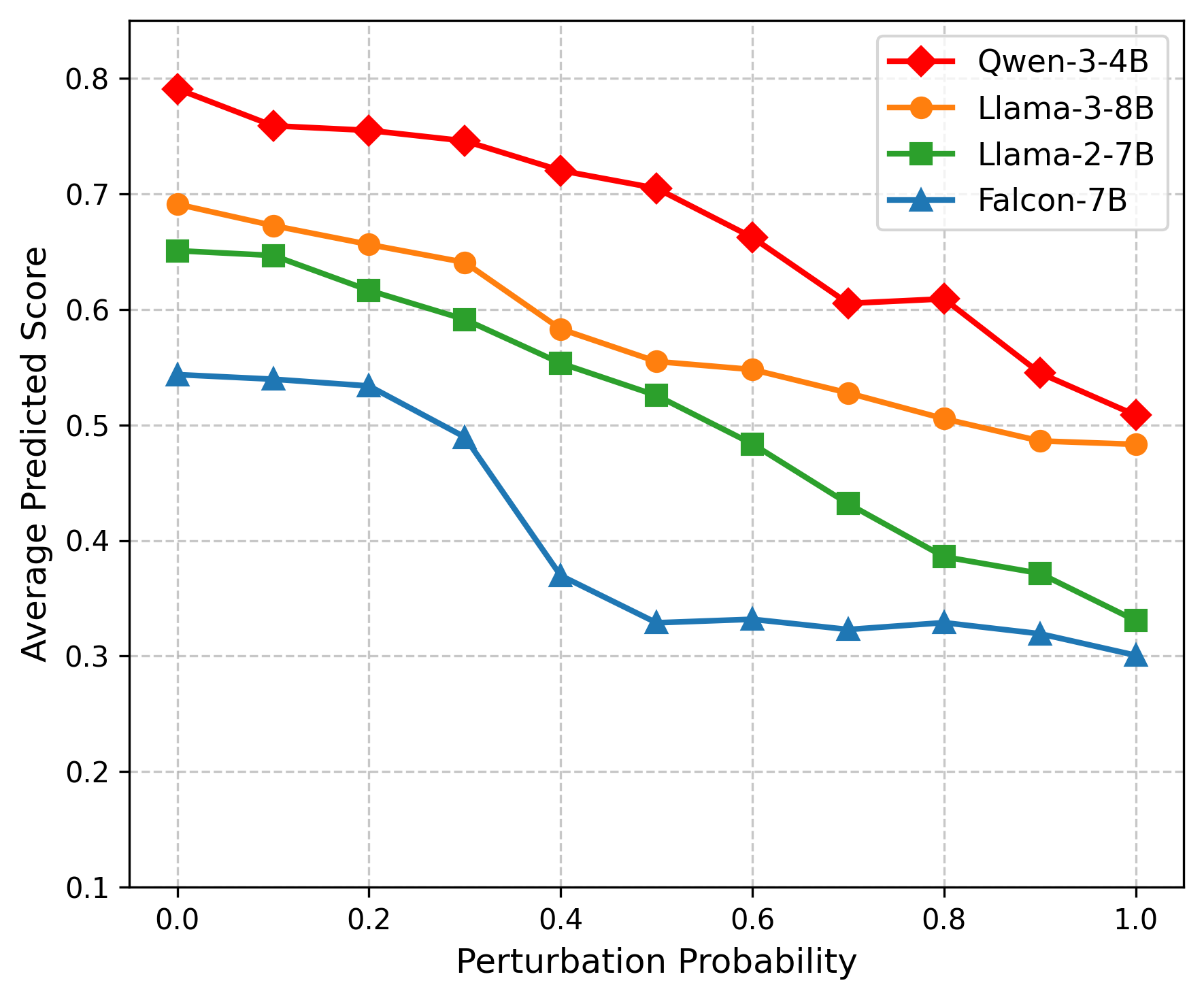}
    \caption{Lyapunov Probe}
    \label{fig:layer:2}
  \end{subfigure}
  
  \caption{Verification of the Lyapunov property across 4 models. Compared with previous probes, the outputs of our Lyapunov probe decrease monotonically with increasing perturbations, enabling more distinct differentiation of hallucinations.}
  \label{fig:lyapunov_verification}
\end{figure}

A fundamental question is whether our trained probes genuinely exhibit Lyapunov stability characteristics or merely learn discriminative patterns from training data. To address this, we evaluate the monotonicity of probe confidence under increasing perturbations. Figure~\ref{fig:lyapunov_verification} compares average factual prediction scores against perturbation magnitudes (ranging from 0.0 to 1.0) for baseline probes versus our Lyapunov Probes.

Baseline probes exhibit erratic, non-monotonic behavior—for instance, Llama-3-8B shows unexpected mid-range fluctuations, indicating these models fail to capture the underlying stability structure. In stark contrast, our Lyapunov Probes demonstrate smooth, monotonic decay across all architectures: Qwen-3-4B decreases from 0.80 to 0.50, Llama-3-8B from 0.69 to 0.48, with similar consistent patterns in other models. This behavior directly satisfies the Lyapunov condition $\frac{\partial V(h, \delta)}{\partial |\delta|} < 0$ formalized in Section~\ref{sec:probe}, confirming that our approach successfully encodes stability principles rather than merely fitting surface patterns. The consistent monotonicity across diverse architectures validates the theoretical foundation of our method.

\subsection{Cross-Domain Generalization Analysis}

A critical advantage of our stability-based framework is its ability to capture universal knowledge boundary properties rather than dataset-specific artifacts. We evaluate this by training probes on TriviaQA and testing on the distinctly different CoQA (dialogue-based) and PopQA (entity-focused) benchmarks. 

Table~\ref{tab:cross_domain_transfer_full} reveals compelling evidence of generalization: our cross-domain probes achieve 20-30 percentage point improvements over probability-based baselines, with Qwen-3-4B reaching 73.12 AUPRC on CoQA and maintaining 48.73 on PopQA despite never seeing similar data during training. More significantly, the performance gap between cross-domain and in-domain probes remains modest—only 5-16 percentage points—suggesting our method captures fundamental stability characteristics that transfer across diverse question types, contexts, and knowledge domains. This transferability is particularly valuable for practical deployment, where collecting labeled data for every domain is prohibitively expensive. The strong cross-domain performance validates our hypothesis that instability at knowledge boundaries follows consistent patterns regardless of surface-level task variations.

\begin{figure}[t]
  \centering
  \includegraphics[width=\linewidth]{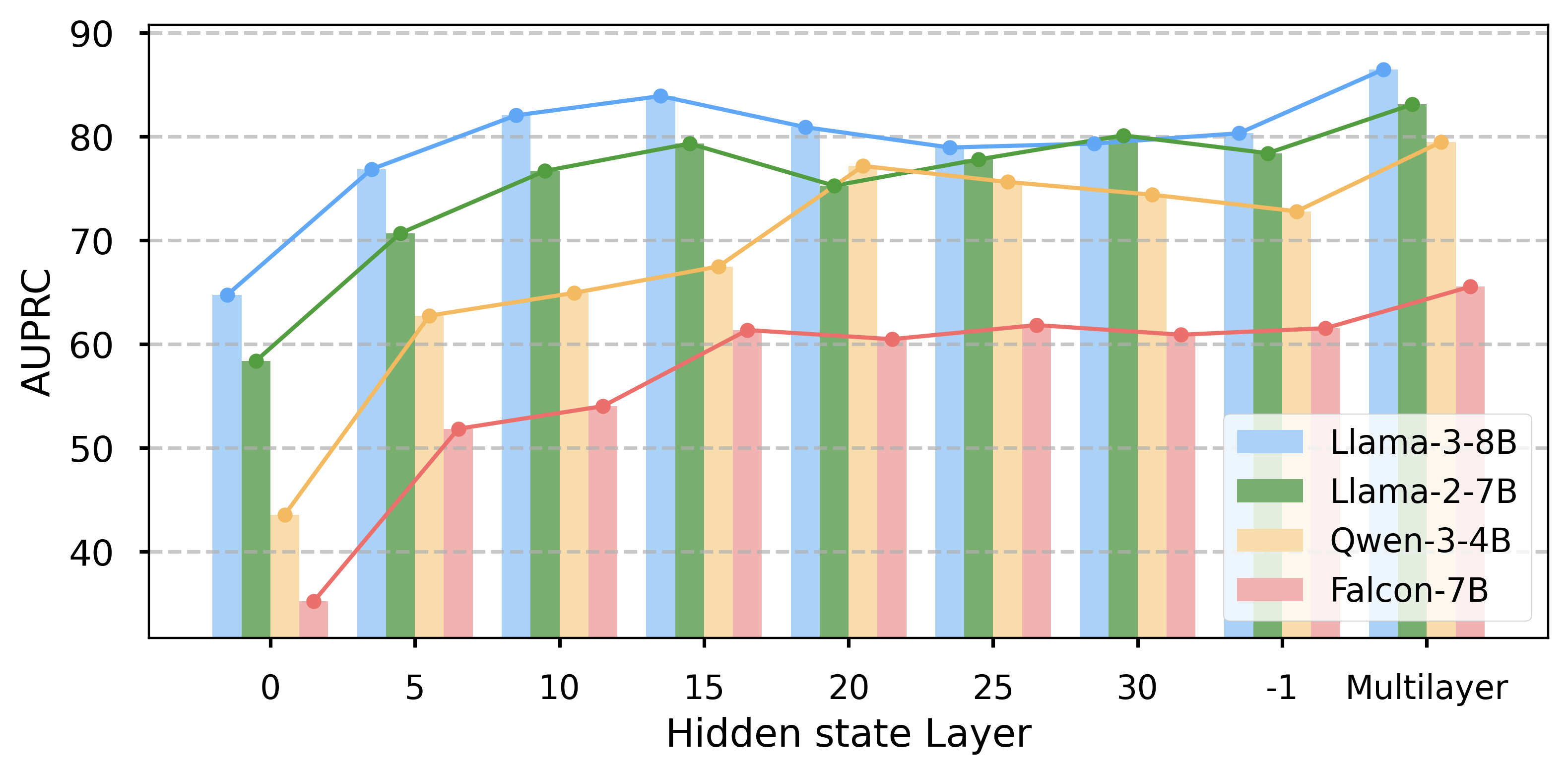}
  \caption{
     Performance comparison among layers from different parts of the model. The performance exhibits significant fluctuations, while the multi-layer fusion method we propose consistently achieves the highest performance.
  }
  \label{fig:layer}
  \vspace{-6pt}
\end{figure}

\subsection{Multi-Layer Representation Analysis}

Our framework aggregates hidden states from multiple transformer layers to capture distributed stability information. To validate this design, we train single-layer probes at various depths and compare them against our multi-layer approach.
Figure~\ref{fig:layer} reveals that optimal single-layer performance varies significantly by architecture, indicating that different models encode stability signals at different depths based on their design and pretraining objectives. Notably, deeper layers (15-32) mostly outperform early layers (0-5) across all architectures, suggesting that mid-to-late representations capture richer semantic and factual content essential for stability assessment.

Critically, our multi-layer aggregation strategy substantially outperforms even the best single-layer configuration across all models.
These consistent gains of 1.8-4.8 percentage points demonstrate that our approach successfully integrates complementary information: early layers provide syntactic features, middle layers capture factual discriminability, and late layers encode output generation dynamics. By leveraging the full representational trajectory rather than committing to a single depth, our method achieves robustness across diverse architectures without requiring architecture-specific layer selection.

\subsection{Component Contribution Analysis}

Table~\ref{tab:ablation_models} systematically evaluates each component by measuring performance when removed. Removing perturbations causes 2-4 point drops, confirming their importance while demonstrating that probes retain discriminative capacity without explicit perturbation signals. Ablating two-stage training shows minimal impact for most models, indicating optimization flexibility, though Qwen-3-4B shows slight improvement, suggesting architecture-dependent optimal training strategies. Restricting to single-layer representations causes the largest degradation, validating multi-layer aggregation as the most critical architectural component for capturing distributed stability information across network depths. Removing the Lyapunov constraint reduces performance by 3-5 points, confirming that explicit monotonic decay enforcement substantially enhances probe quality beyond standard supervised learning. Notably, no single component causes catastrophic failure when removed, with performance remaining 80-90\% of the full model, demonstrating the complementary and robust nature of our design choices.

\section{Conclusion}

This paper proposes a simple yet effective approach to hallucination detection in (Multimodal) Large Language Models by rethinking the problem through the perspective of dynamical systems stability. Instead of viewing hallucination as a simple classification task, we conceptualize (M)LLMs as systems where factual knowledge resides at stable equilibrium points within the representation space, and hallucinations emerge in transitional regions near instability. To address this, we introduce Lyapunov Probes—lightweight models trained with stability-driven constraints that enforce monotonic confidence decay under controlled input perturbations. Using a systematic perturbation framework and a two-stage training process, these probes effectively differentiate between stable, reliable knowledge and unstable, hallucination-prone regions. Extensive experiments across diverse datasets and architectures demonstrate the robustness and effectiveness of our method.

\section*{Acknowledgements}

This work was supported by the National Natural Science Foundation of China under Grant No. 62441617. 
It was also supported by the Postdoctoral Fellowship Program and China Postdoctoral Science Foundation under Grant No. 2024M764093 and Grant No. BX20250485, the Beijing Natural Science Foundation under Grant No. 4254100, and by Beijing Advanced Innovation Center for Future Blockchain and Privacy Computing. 
It was also supported by the Young Elite Scientists Sponsorship Program of the Beijing High Innovation Plan (NO. 20250860).

{
    \small
    \bibliographystyle{ieeenat_fullname}
    \bibliography{main}

@article{llama,
  title={LLaMA: Open and efficient foundation language models},
  author={Touvron, Hugo and Lavril, Thibaut and Izacard, Gautier and Martinet, Xavier and Lachaux, Marie-Anne and Lacroix, Timoth{\'e}e and Rozi{\`e}re, Baptiste and Goyal, Naman and Hambro, Eric and Azhar, Faisal and others},
  journal={arXiv preprint arXiv:2302.13971},
  year={2023}
}

@article{touvron2023llama2,
  title={Llama 2: Open foundation and fine-tuned chat models},
  author={Touvron, Hugo and Martin, Louis and Stone, Kevin and Albert, Peter and Almahairi, Amjad and Babaei, Yasmine and Bashlykov, Nikolay and Batra, Soumya and Bhargava, Prajjwal and Bhosale, Shruti and others},
  journal={arXiv preprint arXiv:2307.09288},
  year={2023}
}

@article{grattafiori2024llama3,
  title={The llama 3 herd of models},
  author={Grattafiori, Aaron and Dubey, Abhimanyu and Jauhri, Abhinav and Pandey, Abhinav and Kadian, Abhishek and Al-Dahle, Ahmad and Letman, Aiesha and Mathur, Akhil and Schelten, Alan and Vaughan, Alex and others},
  journal={arXiv preprint arXiv:2407.21783},
  year={2024}
}

@article{qwen,
  title={Qwen technical report},
  author={Bai, Jinze and Bai, Shuai and Chu, Yunfei and Cui, Zeyu and Dang, Kai and Deng, Xiaodong and Fan, Yang and Ge, Wenbin and Han, Yu and Huang, Fei and others},
  journal={arXiv preprint arXiv:2309.16609},
  year={2023}
}

@article{yang2025qwen3,
  title={Qwen3 technical report},
  author={Yang, An and Li, Anfeng and Yang, Baosong and Zhang, Beichen and Hui, Binyuan and Zheng, Bo and Yu, Bowen and Gao, Chang and Huang, Chengen and Lv, Chenxu and others},
  journal={arXiv preprint arXiv:2505.09388},
  year={2025}
}

@article{almazrouei2023falcon,
  title={The falcon series of open language models},
  author={Almazrouei, Ebtesam and Alobeidli, Hamza and Alshamsi, Abdulaziz and Cappelli, Alessandro and Cojocaru, Ruxandra and Debbah, M{\'e}rouane and Goffinet, {\'E}tienne and Hesslow, Daniel and Launay, Julien and Malartic, Quentin and others},
  journal={arXiv preprint arXiv:2311.16867},
  year={2023}
}

@article{guo2025deepseekr1,
  title={Deepseek-r1: Incentivizing reasoning capability in llms via reinforcement learning},
  author={Guo, Daya and Yang, Dejian and Zhang, Haowei and Song, Junxiao and Zhang, Ruoyu and Xu, Runxin and Zhu, Qihao and Ma, Shirong and Wang, Peiyi and Bi, Xiao and others},
  journal={arXiv preprint arXiv:2501.12948},
  year={2025}
}

@article{liu2024deepseekv3,
  title={Deepseek-v3 technical report},
  author={Liu, Aixin and Feng, Bei and Xue, Bing and Wang, Bingxuan and Wu, Bochao and Lu, Chengda and Zhao, Chenggang and Deng, Chengqi and Zhang, Chenyu and Ruan, Chong and others},
  journal={arXiv preprint arXiv:2412.19437},
  year={2024}
}

@article{claude,
  title={The {Claude} 3 model family: {Opus}, {Sonnet}, {Haiku}},
  author={Anthropic},
  year={2024}
}

@article{gpt4,
  title={{GPT-4} technical report},
  author={Achiam, Josh and Adler, Steven and Agarwal, Sandhini and Ahmad, Lama and Akkaya, Ilge and Aleman, Florencia Leoni and Almeida, Diogo and Altenschmidt, Janko and Altman, Sam and Anadkat, Shyamal and others},
  journal={arXiv preprint arXiv:2303.08774},
  year={2023}
}

@article{joshi2017triviaqa,
  title={Triviaqa: A large scale distantly supervised challenge dataset for reading comprehension},
  author={Joshi, Mandar and Choi, Eunsol and Weld, Daniel S and Zettlemoyer, Luke},
  journal={arXiv preprint arXiv:1705.03551},
  year={2017}
}

@inproceedings{popqa,
  title={When not to trust language models: Investigating effectiveness of parametric and non-parametric memories},
  author={Mallen, Alex and Asai, Akari and Zhong, Victor and Das, Rajarshi and Khashabi, Daniel and Hajishirzi, Hannaneh},
  booktitle={Proceedings of the 61st Annual Meeting of the Association for Computational Linguistics (Volume 1: Long Papers)},
  pages={9802--9822},
  year={2023}
}

@article{reddy2019coqa,
  title={Coqa: A conversational question answering challenge},
  author={Reddy, Siva and Chen, Danqi and Manning, Christopher D},
  journal={Transactions of the Association for Computational Linguistics},
  volume={7},
  pages={249--266},
  year={2019},
  publisher={MIT Press One Rogers Street, Cambridge, MA 02142-1209, USA journals-info~…}
}

@article{mmlu,
  title={Measuring massive multitask language understanding},
  author={Hendrycks, Dan and Burns, Collin and Basart, Steven and Zou, Andy and Mazeika, Mantas and Song, Dawn and Steinhardt, Jacob},
  journal={arXiv preprint arXiv:2009.03300},
  year={2020}
}

@article{mahaut2024factual,
  title={Factual confidence of llms: on reliability and robustness of current estimators},
  author={Mahaut, Mat{\'e}o and Aina, Laura and Czarnowska, Paula and Hardalov, Momchil and M{\"u}ller, Thomas and M{\`a}rquez, Llu{\'\i}s},
  journal={arXiv preprint arXiv:2406.13415},
  year={2024}
}

@article{azaria2023internal,
  title={The internal state of an LLM knows when it's lying},
  author={Azaria, Amos and Mitchell, Tom},
  journal={arXiv preprint arXiv:2304.13734},
  year={2023}
}

@article{kadavath2022language,
  title={Language models (mostly) know what they know},
  author={Kadavath, Saurav and Conerly, Tom and Askell, Amanda and Henighan, Tom and Drain, Dawn and Perez, Ethan and Schiefer, Nicholas and Hatfield-Dodds, Zac and DasSarma, Nova and Tran-Johnson, Eli and others},
  journal={arXiv preprint arXiv:2207.05221},
  year={2022}
}

@article{burns2022discovering,
  title={Discovering latent knowledge in language models without supervision},
  author={Burns, Collin and Ye, Haotian and Klein, Dan and Steinhardt, Jacob},
  journal={arXiv preprint arXiv:2212.03827},
  year={2022}
}

@article{jiang2021can,
  title={How can we know when language models know? on the calibration of language models for question answering},
  author={Jiang, Zhengbao and Araki, Jun and Ding, Haibo and Neubig, Graham},
  journal={Transactions of the Association for Computational Linguistics},
  volume={9},
  pages={962--977},
  year={2021},
  publisher={MIT Press One Rogers Street, Cambridge, MA 02142-1209, USA journals-info~…}
}

@article{yin2023large,
  title={Do large language models know what they don't know?},
  author={Yin, Zhangyue and Sun, Qiushi and Guo, Qipeng and Wu, Jiawen and Qiu, Xipeng and Huang, Xuanjing},
  journal={arXiv preprint arXiv:2305.18153},
  year={2023}
}

@inproceedings{guo2017calibration,
  title={On calibration of modern neural networks},
  author={Guo, Chuan and Pleiss, Geoff and Sun, Yu and Weinberger, Kilian Q},
  booktitle={International conference on machine learning},
  pages={1321--1330},
  year={2017},
  organization={PMLR}
}

@article{fomicheva2020unsupervised,
  title={Unsupervised quality estimation for neural machine translation},
  author={Fomicheva, Marina and Sun, Shuo and Yankovskaya, Lisa and Blain, Fr{\'e}d{\'e}ric and Guzm{\'a}n, Francisco and Fishel, Mark and Aletras, Nikolaos and Chaudhary, Vishrav and Specia, Lucia},
  journal={Transactions of the Association for Computational Linguistics},
  volume={8},
  pages={539--555},
  year={2020},
  publisher={MIT Press One Rogers Street, Cambridge, MA 02142-1209, USA journals-info~…}
}

@inproceedings{gal2016dropout,
  title={Dropout as a bayesian approximation: Representing model uncertainty in deep learning},
  author={Gal, Yarin and Ghahramani, Zoubin},
  booktitle={international conference on machine learning},
  pages={1050--1059},
  year={2016},
  organization={PMLR}
}

@article{lin2022teaching,
  title={Teaching models to express their uncertainty in words},
  author={Lin, Stephanie and Hilton, Jacob and Evans, Owain},
  journal={arXiv preprint arXiv:2205.14334},
  year={2022}
}

@article{tian2023just,
  title={Just ask for calibration: Strategies for eliciting calibrated confidence scores from language models fine-tuned with human feedback},
  author={Tian, Katherine and Mitchell, Eric and Zhou, Allan and Sharma, Archit and Rafailov, Rafael and Yao, Huaxiu and Finn, Chelsea and Manning, Christopher D},
  journal={arXiv preprint arXiv:2305.14975},
  year={2023}
}

@article{xiong2023can,
  title={Can llms express their uncertainty? an empirical evaluation of confidence elicitation in llms},
  author={Xiong, Miao and Hu, Zhiyuan and Lu, Xinyang and Li, Yifei and Fu, Jie and He, Junxian and Hooi, Bryan},
  journal={arXiv preprint arXiv:2306.13063},
  year={2023}
}

@article{wang2022self,
  title={Self-consistency improves chain of thought reasoning in language models},
  author={Wang, Xuezhi and Wei, Jason and Schuurmans, Dale and Le, Quoc and Chi, Ed and Narang, Sharan and Chowdhery, Aakanksha and Zhou, Denny},
  journal={arXiv preprint arXiv:2203.11171},
  year={2022}
}

@inproceedings{manakul2023selfcheckgpt,
  title={Selfcheckgpt: Zero-resource black-box hallucination detection for generative large language models},
  author={Manakul, Potsawee and Liusie, Adian and Gales, Mark},
  booktitle={Proceedings of the 2023 conference on empirical methods in natural language processing},
  pages={9004--9017},
  year={2023}
}

@article{kuhn2023semantic,
  title={Semantic uncertainty: Linguistic invariances for uncertainty estimation in natural language generation},
  author={Kuhn, Lorenz and Gal, Yarin and Farquhar, Sebastian},
  journal={arXiv preprint arXiv:2302.09664},
  year={2023}
}

@article{chen2024inside,
  title={INSIDE: LLMs' internal states retain the power of hallucination detection},
  author={Chen, Chao and Liu, Kai and Chen, Ze and Gu, Yi and Wu, Yue and Tao, Mingyuan and Fu, Zhihang and Ye, Jieping},
  journal={arXiv preprint arXiv:2402.03744},
  year={2024}
}

@article{du2024haloscope,
  title={Haloscope: Harnessing unlabeled llm generations for hallucination detection},
  author={Du, Xuefeng and Xiao, Chaowei and Li, Sharon},
  journal={Advances in Neural Information Processing Systems},
  volume={37},
  pages={102948--102972},
  year={2024}
}

@article{polygraph,
  title={Benchmarking uncertainty quantification methods for large language models with lm-polygraph},
  author={Vashurin, Roman and Fadeeva, Ekaterina and Vazhentsev, Artem and Rvanova, Lyudmila and Vasilev, Daniil and Tsvigun, Akim and Petrakov, Sergey and Xing, Rui and Sadallah, Abdelrahman and Grishchenkov, Kirill and others},
  journal={Transactions of the Association for Computational Linguistics},
  volume={13},
  pages={220--248},
  year={2025},
  publisher={MIT Press 255 Main Street, 9th Floor, Cambridge, Massachusetts 02142, USA~…}
}

@article{lyapunov1992general,
  title={The general problem of the stability of motion},
  author={Lyapunov, Aleksandr Mikhailovich},
  journal={International journal of control},
  volume={55},
  number={3},
  pages={531--534},
  year={1992},
  publisher={Taylor \& Francis}
}

@inproceedings{rodriguez2022lyanet,
  title={Lyanet: A lyapunov framework for training neural odes},
  author={Rodriguez, Ivan Dario Jimenez and Ames, Aaron and Yue, Yisong},
  booktitle={International conference on machine learning},
  pages={18687--18703},
  year={2022},
  organization={PMLR}
}

@article{ode,
  title={Neural ordinary differential equations},
  author={Chen, Ricky TQ and Rubanova, Yulia and Bettencourt, Jesse and Duvenaud, David K},
  journal={Advances in neural information processing systems},
  volume={31},
  year={2018}
}

@inproceedings{cipriani2024minimax,
  title={A minimax optimal control approach for robust neural ODEs},
  author={Cipriani, Cristina and Scagliotti, Alessandro and W{\"o}hrer, Tobias},
  booktitle={2024 European Control Conference (ECC)},
  pages={58--64},
  year={2024},
  organization={IEEE}
}

@article{westny2023stability,
  title={Stability-informed initialization of neural ordinary differential equations},
  author={Westny, Theodor and Mohammadi, Arman and Jung, Daniel and Frisk, Erik},
  journal={arXiv preprint arXiv:2311.15890},
  year={2023}
}

@article{llava,
  title={Visual instruction tuning},
  author={Liu, Haotian and Li, Chunyuan and Wu, Qingyang and Lee, Yong Jae},
  journal={Proceedings of the Advances in Neural Information Processing Systems},
  volume={36},
  year={2024}
}

@inproceedings{llava1.5,
  title={Improved baselines with visual instruction tuning},
  author={Liu, Haotian and Li, Chunyuan and Li, Yuheng and Lee, Yong Jae},
  booktitle={Proceedings of the IEEE/CVF Conference on Computer Vision and Pattern Recognition},
  pages={26296--26306},
  year={2024}
}

@article{bai2023qwenvl,
  title={Qwen-VL: A versatile vision-language model for understanding, localization, text reading, and beyond},
  author={Bai, Jinze and Bai, Shuai and Yang, Shusheng and Wang, Shijie and Tan, Sinan and Wang, Peng and Lin, Junyang and Zhou, Chang and Zhou, Jingren},
  journal={arXiv preprint arXiv:2308.12966},
  volume={1},
  number={2},
  pages={3},
  year={2023}
}

@article{wang2024qwenvl2,
  title={Qwen2-vl: Enhancing vision-language model's perception of the world at any resolution},
  author={Wang, Peng and Bai, Shuai and Tan, Sinan and Wang, Shijie and Fan, Zhihao and Bai, Jinze and Chen, Keqin and Liu, Xuejing and Wang, Jialin and Ge, Wenbin and others},
  journal={arXiv preprint arXiv:2409.12191},
  year={2024}
}

@article{bai2025qwenvl2.5,
  title={Qwen2. 5-vl technical report},
  author={Bai, Shuai and Chen, Keqin and Liu, Xuejing and Wang, Jialin and Ge, Wenbin and Song, Sibo and Dang, Kai and Wang, Peng and Wang, Shijie and Tang, Jun and others},
  journal={arXiv preprint arXiv:2502.13923},
  year={2025}
}

@article{internvl1_5,
  title={How far are we to GPT-4V? Closing the gap to commercial multimodal models with open-source suites},
  author={Chen, Zhe and Wang, Weiyun and Tian, Hao and Ye, Shenglong and Gao, Zhangwei and Cui, Erfei and Tong, Wenwen and Hu, Kongzhi and Luo, Jiapeng and Ma, Zheng and others},
  journal={arXiv preprint arXiv:2404.16821},
  year={2024}
}

@article{chiang2023vicuna,
  title={Vicuna: An open-source chatbot impressing GPT-4 with 90\%* ChatGPT quality},
  author={Chiang, Wei-Lin and Li, Zhuohan and Lin, Zi and Sheng, Ying and Wu, Zhanghao and Zhang, Hao and Zheng, Lianmin and Zhuang, Siyuan and Zhuang, Yonghao and Gonzalez, Joseph E and others},
  journal={See https://vicuna. lmsys. org},
  volume={2},
  number={3},
  pages={6},
  year={2023}
}

@article{gpt4v,
  title={The dawn of {LMMs}: Preliminary explorations with {GPT-4V(ision)}},
  author={Yang, Zhengyuan and Li, Linjie and Lin, Kevin and Wang, Jianfeng and Lin, Chung-Ching and Liu, Zicheng and Wang, Lijuan},
  journal={arXiv preprint arXiv:2309.17421},
  volume={9},
  number={1},
  pages={1},
  year={2023}
}

@article{textcot,
  title={Textcot: Zoom in for enhanced multimodal text-rich image understanding},
  author={Luan, Bozhi and Feng, Hao and Chen, Hong and Wang, Yonghui and Zhou, Wengang and Li, Houqiang},
  journal={arXiv preprint arXiv:2404.09797},
  year={2024}
}

@inproceedings{ding2023improving,
  title={Improving handwritten OCR with training samples generated by glyph conditional denoising diffusion probabilistic model},
  author={Ding, Haisong and Luan, Bozhi and Gui, Dongnan and Chen, Kai and Huo, Qiang},
  booktitle={International Conference on Document Analysis and Recognition},
  pages={20--37},
  year={2023},
  organization={Springer}
}

@article{adaptprune,
  title={Multi-Cue Adaptive Visual Token Pruning for Large Vision-Language Models},
  author={Luan, Bozhi and Zhou, Wengang and Feng, Hao and Wang, Zhe and Li, Xiaosong and Li, Houqiang},
  journal={arXiv preprint arXiv:2503.08019},
  year={2025}
}

@inproceedings{textvqa,
  title={Towards VQA models that can read},
  author={Singh, Amanpreet and Natarajan, Vivek and Shah, Meet and Jiang, Yu and Chen, Xinlei and Batra, Dhruv and Parikh, Devi and Rohrbach, Marcus},
  booktitle={Proceedings of the IEEE/CVF Conference on Computer Vision and Pattern Recognition},
  pages={8317--8326},
  year={2019}
}

@inproceedings{gurari2018vizwiz,
  title={VizWiz grand challenge: Answering visual questions from blind people},
  author={Gurari, Danna and Li, Qing and Stangl, Abigale J and Guo, Anhong and Lin, Chi and Grauman, Kristen and Luo, Jiebo and Bigham, Jeffrey P},
  booktitle={Proceedings of the IEEE/CVF Conference on Computer Vision and Pattern Recognition},
  pages={3608--3617},
  year={2018}
}

@article{pope,
  title={Evaluating object hallucination in large vision-language models},
  author={Li, Yifan and Du, Yifan and Zhou, Kun and Wang, Jinpeng and Zhao, Wayne Xin and Wen, Ji-Rong},
  journal={arXiv preprint arXiv:2305.10355},
  year={2023}
}

@misc{mmebenchmark,
      title={MME: A Comprehensive Evaluation Benchmark for Multimodal Large Language Models}, 
      author={Chaoyou Fu and Peixian Chen and Yunhang Shen and Yulei Qin and Mengdan Zhang and Xu Lin and Jinrui Yang and Xiawu Zheng and Ke Li and Xing Sun and Yunsheng Wu and Rongrong Ji and Caifeng Shan and Ran He},
      year={2025},
      eprint={2306.13394},
      archivePrefix={arXiv},
      primaryClass={cs.CV},
      url={https://arxiv.org/abs/2306.13394}, 
}

@inproceedings{lmmseval,
  title={Lmms-eval: Reality check on the evaluation of large multimodal models},
  author={Zhang, Kaichen and Li, Bo and Zhang, Peiyuan and Pu, Fanyi and Cahyono, Joshua Adrian and Hu, Kairui and Liu, Shuai and Zhang, Yuanhan and Yang, Jingkang and Li, Chunyuan and others},
  booktitle={Findings of the Association for Computational Linguistics: NAACL 2025},
  pages={881--916},
  year={2025}
}

@article{mllmhallucination,
  title={Hallucination of multimodal large language models: A survey},
  author={Bai, Zechen and Wang, Pichao and Xiao, Tianjun and He, Tong and Han, Zongbo and Zhang, Zheng and Shou, Mike Zheng},
  journal={arXiv preprint arXiv:2404.18930},
  year={2024}
}
}

\end{document}